 \documentclass[runningheads, envcountsame, a4paper]{llncs}
\usepackage{booktabs}
\usepackage{multirow}
\usepackage{afterpage}
\usepackage{cite}
\usepackage{graphicx}
\usepackage[table]{xcolor}
\usepackage[hyphens]{url}
\usepackage[misc]{ifsym}

\begin{document}
\title{Transformers: ``The End of History''\\ for Natural Language Processing?}
%
%
\author{Anton Chernyavskiy\inst{1}{\Letter} \and
Dmitry Ilvovsky\inst{1} \and
Preslav Nakov\inst{2}}

\authorrunning{A. Chernyavskiy et al.}
%
\institute{HSE University, Russian Federation \\ \email{aschernyavskiy\_1@edu.hse.ru, dilvovsky@hse.ru} \and
Qatar Computing Research Institute, HBKU, Doha, Qatar \\
\email{pnakov@hbku.edu.qa}}
\maketitle              
\setcounter{footnote}{0}
\begin{abstract}
Recent advances in neural architectures, such as the Transformer, coupled with the emergence of large-scale pre-trained models such as BERT, have revolutionized the field of Natural Language Processing (NLP), pushing the state of the art for a number of NLP tasks. A rich family of variations of these models has been proposed, such as RoBERTa, ALBERT, and XLNet, but fundamentally, they all remain limited in their ability to model certain kinds of information, and they cannot cope with certain information sources, which was easy for pre-existing models. Thus, here we aim to shed light on some important theoretical limitations of pre-trained BERT-style models that are inherent in the general Transformer architecture. First, we demonstrate in practice on two general types of tasks---segmentation and segment labeling---and on four datasets that these limitations are indeed harmful and that addressing them, even in some very simple and na\"{i}ve ways, can yield sizable improvements over vanilla RoBERTa and XLNet models.
Then, we offer a more general discussion on desiderata for future additions to the Transformer architecture that would increase its expressiveness, which we hope could help in the design of the next generation of deep NLP architectures.

\end{abstract}

\section{Introduction}
The history of Natural Language Processing (NLP) has seen several stages: first, rule-based, e.g.,~think of the expert systems of the 80s, then came the statistical revolution, and now along came the neural revolution. The latter was enabled by a combination of deep neural architectures, specialized hardware, and the existence of large volumes of data. Yet, the revolution was going slower in NLP compared to other fields such as Computer Vision, which were quickly and deeply transformed by the emergence of large-scale pre-trained models, which were in turn enabled by the emergence of large datasets such as ImageNet. 

Things changed in 2018, when NLP finally got its ``ImageNet moment'' with the invention of BERT \cite{devlin-etal-2019-bert}.\footnote{A notable previous promising attempt was ELMo  \cite{peters-etal-2018-deep}, but it became largely outdated in less than a year.}
This was enabled by recent advances in neural architectures, such as the Transformer \cite{DBLP:journals/corr/VaswaniSPUJGKP17}, followed by the emergence of large-scale pre-trained models such as BERT, which eventually revolutionized NLP and pushed the state of the art for a number of NLP tasks. A rich family of variations of these models have been proposed, such as RoBERTa \cite{DBLP:journals/corr/abs-1907-11692}, ALBERT \cite{lan2019albert}, and XLNet \cite{NIPS2019_8812}. For some researchers, it felt like this might very well be the ``End of History'' for NLP (\`a la Fukuyama\footnote{\url{http://en.wikipedia.org/wiki/The_End_of_History_and_the_Last_Man}}). 

It was not too long before researchers started realizing that BERT and Transformer architectures in general, despite their phenomenal success, remained fundamentally limited in their ability to model certain kinds of information, which was natural and simple for the old-fashioned feature-based models. Although BERT does encode some syntax, semantic, and linguistic features, it may not use them in downstream tasks \cite{kovaleva-etal-2019-revealing}.
It ignores negation \cite{Ettinger_2020}, and it might need to be combined with Conditional Random Fields (CRF) to improve its performance for some tasks and languages, most notably for sequence classification tasks \cite{souza2019portuguese}.
There is a range of sequence tagging tasks where entities have different lengths (not 1-3 words as in the classical named entity  recognition formulation), and sometimes their continuity is required, e.g.,~for tagging in court papers. Moreover, in some problem formulations, it is important to accurately process the boundaries of the spans (in particular, the punctuation symbols), which turns out to be something that Transformers are not particularly good at (as we will discuss below). 

In many sequence classification tasks, some classes are described by specific features. Besides, a very large contextual window may be required for the correct classification, which is a problem for Transformers because of the quadratic complexity of calculating their attention weights.\footnote{Some solutions were proposed such as Longformer \cite{Beltagy2020Longformer}, Performer~\cite{DBLP:journals/corr/abs-2009-14794}, Linformer~\cite{DBLP:journals/corr/abs-2006-04768}, Linear Transformer~\cite{DBLP:journals/corr/abs-2006-16236}, and Big Bird~\cite{DBLP:journals/corr/abs-2007-14062}.}

Is it possible to guarantee that BERT-style models will carefully analyze all these cases? This is what we aim to explore below. Our contributions can be summarized as follows:

\begin{itemize}
    \item We explore some theoretical limitations of pre-trained BERT-style models when applied to sequence segmentation and labeling tasks. We argue that these limitations are not limitations of a specific model, but stem from the general Transformer architecture.
    \item We demonstrate in practice on two different tasks (one on segmentation, and one on segment labeling) and on four datasets that it is possible to improve over state-of-the-art models such as BERT, RoBERTa, XLNet, and this can be achieved with simple and na\"{i}ve approaches, such as feature engineering and post-processing.
    \item Finally, we propose desiderata for attributes to add to the Transformer architecture in order to increase its expressiveness, which could guide the design of the next generation of deep NLP architectures.
\end{itemize}

The rest of our paper is structured as follows. Section \ref{sec:2}~summarizes related prior research. Section \ref{sec:3}~describes the tasks we address. Section~\ref{sec:4} presents the models and the modifications thereof. Section~\ref{sec:5} outlines the experimental setup. Section~\ref{sec:6} describes the experiments and the evaluation results. Section~\ref{sec:7} provides key points that lead to further general potential improvements of Transformers. Section~\ref{sec:8} concludes and points to possible directions for future work.

\section{Related Work}
\label{sec:2}

\paragraph{Studies of what 
BERT learns and what it can represent} There is a large number of papers that study what kind of information can be learned with BERT-style models and how attention layers capture this information; a survey is presented in \cite{rogers2020primer}. It was shown that BERT learns syntactic features \cite{goldberg2019assessing, Liu_2019}, semantic roles and entities types \cite{tenney2019learn}, linguistic information and subject-verb agreement \cite{jawahar-etal-2019-bert}.
Note that the papers that explore what BERT-style models can encode do not indicate that they directly use such knowledge \cite{kovaleva-etal-2019-revealing}. Instead, we focus on what is \textit{not} modeled, and we explore some general limitations.

\paragraph{Limitations of BERT/Transformer} Indeed, Kovaleva et al. (2019) \cite{kovaleva-etal-2019-revealing} revealed that vertical self-attention patterns generally come from pre-training tasks rather than from task-specific linguistic reasoning and the model is over-parametrized. Ettinger (2020) \cite{Ettinger_2020} demonstrated that BERT encodes some semantics, but is fully insensitive to negation. Sun et al. (2020) \cite{sun2020advbert} showed that BERT-style models are erroneous in simple cases, e.g.,~they do not correctly process word sequences with misspellings. They also have bad representations of floating point numbers for the same tokenization reason \cite{Wallace_2019}. Moreover, it is easy to attack them with adversarial examples \cite{jin2019bert}. Durrani et al. (2019) \cite{durrani-etal-2019-one} showed that BERT subtoken-based representations are better for modeling syntax, while ELMo character-based representations are preferable for modeling morphology. It also should be noticed that hyper-parameter tuning is a very non-trivial task, not only for NLP engineers but also for advanced NLP researchers \cite{Popel_2018}. Most of these limitations are low-level and technical, or are related to a specific architecture (such as BERT). In contrast, we single out the general limitations of the  Transformer at a higher level, but which can be technically confirmed, and provide desiderata for their elimination.

\paragraph{Fixes of BERT/Transformer} Many improvements of the original BERT model have been proposed: RoBERTa (changed the language model masking, the learning rate, the dataset size), DistilBERT \cite{sanh2019distilbert} (distillation to significantly reduce the number of parameters), ALBERT (cross-layer parameter sharing, factorized embedding parametrization), Transformer-XL \cite{Dai_2019} (recurrence mechanism and relative positional encoding to improve sequence modeling), XLNet (permutation language modeling to better model bidirectional relations), BERT-CRF \cite{arkhipov-etal-2019-tuning, souza2019portuguese} (dependencies between the posteriors for structure prediction helped in some tasks and languages), KnowBERT \cite{Peters_2019} (incorporates external knowledge). Most of these models pay attention only to 1--2 concrete fixes, whereas our paper aims at more general Transformer limitations.

\section{Tasks}\label{sec:3}

In this section, we describe two tasks and four datasets that we used for experiments.

\subsection{Propaganda Detection}

We choose the task of Detecting Propaganda Techniques in News Articles (SemEval-2020 Task 11)\footnote{The official task webpage: \url{http://propaganda.qcri.org/semeval2020-task11/}} as the main for experiments. Generally, it is formulated as finding and classifying all propagandistic fragments in the text \cite{DaSanMartinoSemeval20task11}. To do this, two subtasks are proposed: (\emph{i})~\textit{span identification (SI)}, i.e.,~selection of all propaganda spans within the article, 
(\emph{ii})~\textit{technique classification (TC)}, i.e.,~multi-label classification of each span into 14 classes.
The corpus with a detailed description of propaganda techniques is presented in \cite{da-san-martino-etal-2019-fine}.

The motivation for choosing this task is triggered by several factors. First, two technically different problems are considered, which can be formulated at a general level (multi-label sequence classification and binary token labeling). Second, this task has specificity necessary for our research, unlike standard named entity recognition. Thus, traditional NLP methods can be applied over the set of hand-crafted features: sentiment, readability scores, length, etc. Here, length is a strong feature due to the data statistics \cite{da-san-martino-etal-2019-fine}. Moreover, spans can be nested, while span borders vary widely and may include punctuation symbols. Moreover, sometimes Transformer-based models face the problem of limited input sequence length. In this task, such a  problem appears with the classification of ``Repetition'' spans. By definition, this class includes spans that have an intentional repetition of the same information. This information can be repeated both in the same sentence and in very distant parts of the text.

\subsection{Keyphrase Extraction}

In order to demonstrate the transferability of the studied limitations between datasets, we further experimented with the task of Extracting Keyphrases and Relations from Scientific Publications, using the dataset from SemEval-2017 Task 10 \cite{augenstein-etal-2017-semeval}. We focus on the following two subtasks: (\emph{i})~\textit{keyphrase identification (KI)}, i.e.,~search of all keyphrases within the text, (\emph{ii})~~\textit{keyphrase classification (KC)}, i.e.,~multi-class classification of given keyphrases into three classes. According to the data statistics, the length of the phrases is a strong feature. Also, phrases can be nested inside one other, and many of them are repeated across different articles. So, these subtasks allow us to demonstrate a number of issues.

\section{Method}
\label{sec:4}

Initially, we selected the most successful approach as the baseline from Transformer-based models (BERT, RoBERTa, ALBERT, and XLNet). In both propaganda detection subtasks, it turned out to be RoBERTa, which is an optimized version of the standard BERT with a modified pre-training procedure. Whereas in both keyphrase extraction subtasks, it turned out to be XLNet, which is a language model that aims to better study bidirectional links or relationships in a sequence of words. From a theoretical point of view, investigated results and researched problems should be typical for other Transformer-based models, such as BERT and DistilBERT. Nonetheless, we additionally conduct experiments with both XLNet and RoBERTa for both tasks for a better demonstration of the universality of our findings.

\subsection{Token Classification}

We reformulate the SI and the KI tasks as ``named entity recognition'' tasks. Specifically, in the SI task, for each span, all of its internal tokens are assigned to the ``\texttt{PROP}'' class and the rest to ``\texttt{O}'' (\textbf{O}utside). Thus, this is a binary token classification task. At the same time, various types of encoding formats are studied. Except for the above described Inside-Outside classification, we further consider BIO (\textbf{B}egin) and BIEOS (\textbf{B}egin, \textbf{E}nd, \textbf{S}ingle are added) tags encodings. Such markups theoretically can provide better processing for border tokens \cite{Ratinov2009DesignCA}.

In order to ensure the sustainability of the trained models, we create an ensemble of three models trained with the same hyper-parameters, but using different random seeds. We merge the intersecting spans during the ensemble procedure (intervals union).

\paragraph{End-to-End Training with CRFs}
Conditional Random Fields (CRF) \cite{10.5555/645530.655813} can qualitatively track the dependencies between the tags in the markup. Therefore, this approach has gained great popularity in solving the problem of extracting named entities with LSTMs or RNNs. Advanced Transformer-based models generally can model relationships between words at a good level due to the attention mechanism, but adding a CRF layer theoretically is not unnecessary. The idea is that we need to model the relationships not only between the input tokens but also between the output labels.

Our preliminary study showed that both RoBERTa and XLNet are make classification errors even when choosing tags in named entity recognition (NER) encodings with clear rules. For example, in the case of BIO, the ``\texttt{I-PROP}'' tag can only go after the ``\texttt{B-PROP}'' tag. However, RoBERTa produced results with a sequence of tags such as ``\texttt{O-PROP} \texttt{I-PROP} \texttt{O-PROP}'' for some inputs. Here, it is hard to determine where the error was, but the CRF handles such cases from a probabilistic point of view.
We use the CRF layer instead of the standard model classification head to apply the end-to-end training. Here, we model connections only between neighboring subtokens since our main goal is the proper sequence analysis. Thus, the subtokens that are not placed at the beginning of words are ignored (i.e.,~of the format \textit{\#\#smth}).

\paragraph{RoBERTa, XLNet, and Punctuation Symbols}
In the SI task, there is one more problem that even the CRF layer cannot always handle. It is the processing of punctuation and quotation marks at the span borders. Clark et al. (2019) \cite{DBLP:journals/corr/abs-1906-04341} and Kovaleva et al. (2019) \cite{kovaleva-etal-2019-revealing} showed that BERT generally has high token--token attention to the \texttt{[SEP]} token, to the periods, and to the commas, as they are the most frequent tokens. However, we found out that large attention weights to punctuation may still not be enough for some tasks.

In fact, a simple rule can be formulated to address this problem: a span cannot begin or end with a punctuation symbol, unless it is enclosed in quotation marks. With this observation in mind, we apply post-processing of the spans borders by adding missing quotation marks and also by filtering punctuation symbols in case they were absent.

\subsection{Sequence Classification}

We model the TC task in the same way as the KC task, that is, as a multi-class sequence classification problem. We create a fairly strong baseline to achieve better results. First, the context is used, since spans in both tasks can have various meanings in different contexts. Thus, we select the entire sentence that contains the span for this purpose. In this case, we consider two possible options: (\emph{i})~highlight the span with the special limiting tokens and submit to the model only one input; (\emph{ii})~make two inputs: one for the span and one for the context. Moreover, to provide a better initialization of the model and to share some additional knowledge from other data in the TC task, we apply the transfer learning strategy from the SI task.

Just like in the token classification problem, we compose an ensemble of models for the same architecture, but with three different random seed initializations. We do this in order to stabilize the model, and this is not a typical ensemble of different models.

\paragraph{Input Length}
BERT does not have a mechanism to perform explicit character/word/subword counting. Exactly this problem and the lack of good consistency between the predicted tags may cause a problem with punctuation (quotation marks) in the sequence tagging task, since BERT theoretically cannot accurately account for the number of opening/closing quotation marks (as it cannot count).

\begin{figure}[tbh]
\centering
\includegraphics[scale=0.85]{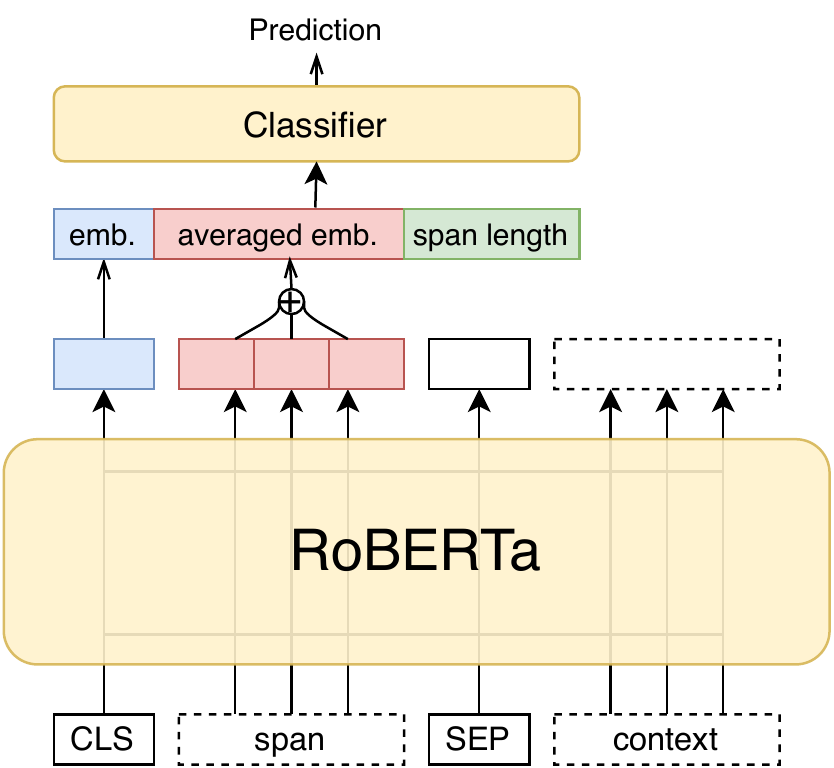}
\caption{The RoBERTa model takes an input span and the context (sentence with the span). It combines the embedding of the \texttt{[CLS]} token, the averaged embedding of all span tokens, and the span length as a feature.}
\label{fig_2}
\end{figure}

In order to explicitly take into account the input sequence size in the model, we add a length feature to the \texttt{[CLS]} token embedding, as it should contain all the necessary information to solve the task (see Figure~\ref{fig_2}). It may be also useful to pre-process the length feature through binning. In this case, it is possible to additionally create trainable embeddings associated with each bin or directly to add an external knowledge from a gazetteer containing relevant information about the dataset according to the given bin (we will consider gazetteers below).

In addition to the input length in characters (or in tokens), it may be useful to add other quantitative features such as the number of question or exclamation symbols.

\paragraph{Span Embeddings}
In the TC task, we concatenate the \texttt{[CLS]} token representation with the span embedding obtained by averaging all token embeddings from the last layer to submit to the classifier (end-to-end training). Note that the added embedding contains information about the degree of propaganda in the classified span as an initialization, since we transfer a model from another task. Moreover, this model can reconfigure it to serve other features during the training process. Also, it may be useful to join embeddings obtained by the max-pool operation or taken from other layers.

\paragraph{Training a Hand-Crafted Gazetteer}
Gazetteers can provide external relevant information about entities in NER tasks. As some propaganda techniques are often described by the same words, it might be a good idea to construct and to use a gazetteer of words for each technique. While in NER, gazetteers are externally constructed to provide additional knowledge, here we use the training data to construct our gazetteer. We create a hash map, where the keys are spans pre-processed by the Porter stemmer \cite{porter1980algorithm}, and the values are distributions of the classes in which spans are present in the training dataset.

There are several ways to use this gazetteer. First, we can use these frequency representations as additional features and concatenate them with the \texttt{[CLS]} token in the same way as described for the length and the span embedding. However, in this case, over-fitting may occur since such a feature will contain a correct label. The second method is based on post-processing. The idea is to increase the probability of each class of spans by some value (e.g.,~+0.5) if the span of this class is present in the gazetteer.

\paragraph{Class Insertions}
Earlier, we described the problem of non-perfect spatially consistent class predictions for the token labeling task. For the sequence classification task, it may be expressed as the incorrect nesting of classes. That is, the model can produce a markup in which the span of class \texttt{A} is nested in the span of another class \texttt{B}, but there are no such cases in the training data. If we believe that the training set gives us an almost complete description of the researched problem, such a classification obviously cannot be correct. 

The simplest solution is again post-processing. One possibility is to choose a pair of spans that have maximal predicted probability and the correct nesting. Another option is to choose a pair of classes with a maximal probability $p(x)p(y)p(A)$. Here, $p(x)$ is the predicted probability that the span has the label $x$, and $p(A)$ is the estimated probability of the nesting case $A$, where a span of class $x$ is inside the span of class $y$. To estimate $p(A)$, we calculate the co-occurrence matrix of nesting classes in the training set, and we apply softmax with temperature $t$ over this matrix to obtain probabilities. The temperature parameter is adjusted for each model on validation. We use the first approach in the TC task. As there are only three classes and all class insertions are possible, we apply the second approach with $t = 0.26$ in the KC task.

\paragraph{Specific Classes: ``Repetition''}
In some cases, the entire text of the input document might be needed as a context (rather than just the current sentence) in order for the model to be able to predict the correct propaganda technique for a given span. This is the case of the \emph{repetition} technique.

As a solution, we apply a special post-processing step. Let $k$ be the number of occurrences of the considered span in the set of spans allocated for prediction within the article and $p$ be the probability of the \emph{repetition} class predicted by the source model. We apply the following formula:

\begin{equation}
\hat{p}= \left\{
  \begin{array}{l}
    1,\hspace{5pt}\textrm{if}\hspace{3pt} k \geq 3\,\,\, \textrm{or}\,\,\, (k=2\,\,\, \textrm{and}\,\,\, p \geq t_1) \\
    0,\hspace{5pt}\textrm{if}\hspace{3pt}  k=1\,\,\, \textrm{and}\, p \leq \,\,t_2 \\
    p,\hspace{5pt}\textrm{otherwise} \\
  \end{array}
\right.
\end{equation}

We use the following values for the probability thresholds: $t_1=0.001$ and $t_2=0.99$. Note that since the repetition may be contained in the span itself, it is incorrect to nullify the probabilities of the unique spans.

\paragraph{Multi-label Classification}
If the same span can have multiple labels, it is necessary to apply supplementary post-processing of the predictions. Thus, if the same span is asked several times during the testing process (the span is determined by its coordinates in the text, and in the TC task, multiple labels are signalled by repeating the same span multiple times in the test set), then we assign different labels to the different instances of that span, namely the top among the most likely predictions.

\section{Experimental Setup}\label{sec:5}

Below, we describe the data we used and the parameter settings for our experiments for all the tasks.

\subsection{Data}

\paragraph{Propaganda Detection} The dataset provided for the SemEval-2020  task 11 contains 371 English articles for training, 75 for development, and 90 for testing. Together, the training and the  testing sets contain 6,129 annotated spans. While there was an original partitioning of the data into training, development, and testing, the latter was only available via the task leaderboard, and was not released. Thus, we additionally randomly split the training data using a 80:20 ratio to obtain new training and validation sets.
The evaluation measure for the SI task is the variant of the F$_1$ measure described in \cite{da-san-martino-etal-2019-fine}: it penalizes for predicting too long or too short spans (compared to the gold span) and generally correlates with the standard F$_1$ score for tokens. Micro-averaged F$_1$ score is used for the TC task, which is equivalent to accuracy.

\paragraph{Keyphrase Extraction} The dataset provided for SemEval-2017 task 10 contains 350 English documents for training, 50 for development, and 100 for testing. In total, the training and the testing sets contain 9,945 annotated keyphrases. The evaluation measure for both sub-tasks is micro-averaged F$_1$ score.

\subsection{Parameter Setting}

We started with pre-trained model checkpoints and baselines as from the HuggingFace Transformers library,\footnote{\url{http://github.com/huggingface/transformers}} and we implemented our modifications on top of them. We used RoBERTa-large and XLNet-large, as they performed better than their base versions in our preliminary experiments.

We selected hyper-parameters according to the recommendations in the original papers using our validation set and we made about 10--20 runs to find the best configuration. We used grid-search over \{5e-6, 1e-5, 2e-5, 3e-5, 5e-5\} for the optimal learning rate. Thus, we fix the following in the propaganda detection problem: learning rate of 2e-5 (3e-5 for XLNet in the TC task), batch size of 24, maximum sequence length of 128 (128 is fixed as it is long enough to encode the span; besides, there are very few long sentences in our datasets), Adam optimizer with a linear warm-up of 500 steps. The sequence length and the batch size are selected as the maximum possible for our GPU machine (3 GeForce GTX 1080 GPUs). We performed training for 30 epochs with savings every two epochs and we selected the best checkpoints on the validation set (typically, it was 10--20 epochs). We found that uncased models should be used for the SI task, whereas the cased model were better for the TC task.

\bgroup
\begin{table}[tbh]
\begin{center}
\small
\begin{tabular}{c l l}
\toprule
\bf Task & \bf Approach  & \bf F1  \\
\midrule
 \multirow{8}{*}{\shortstack{\bf SI}}  & RoBERTa (BIO encoding) & $46.91$ \\ 
& \hspace{10pt}+ CRF & $48.54_{\bf\,\uparrow 1.63}$ \\
& \hspace{10pt}+ punctuation post-processing & $47.54_{\bf\,\uparrow 0.63}$ \\
& \textit{Overall} & $48.87_{\bf\,\uparrow 1.96}$  \\
\cmidrule{2-3}
& XLNet (BIO encoding) & $46.47 $ \\ 
& \hspace{10pt}+ CRF & $46.68_{\bf\,\uparrow 0.21}$ \\
& \hspace{10pt}+ punctuation post-processing & $46.76_{\bf\,\uparrow 0.29}$ \\
& \textit{Overall} & $47.05_{\bf\,\uparrow 0.58}$\\
\midrule
\multirow{4}{*}{\shortstack{\bf KI}} & RoBERTa (BIO encoding) & $57.85$  \\ 
& \hspace{10pt}+ CRF & $58.59_{\bf\,\uparrow 0.74}$ \\
\cmidrule{2-3}
& XLNet (BIO encoding) & $58.80 $ \\ 
& \hspace{10pt}+ CRF & $60.11_{\bf\,\uparrow 1.31}$ \\
\bottomrule
\end{tabular}
\end{center}
\caption{Analysis of RoBERTa and XLNet modifications for sequential classification tasks: span identification and keyphrase identification. \textit{Overall} is the simultaneous application of two improvements.}
\label{table_1}
\end{table}
\egroup

For keyphrase extraction, for the KI task, we used a learning rate of 2e-5 (and 3e-5 for RoBERTa-CRF), a batch size of 12, a maximum sequence length of 64, Adam optimizer with a linear warm-up of 60 steps. For the KC task, we used a learning rate of 2e-5 (1e-5 for XLNet-Length) 
and a head learning rate of 1e-4 (in cases with the \textit{Length} feature), batch size of 20 (10 for XLNet-Length), maximum sequence length of 128, and the Adam optimizer with a linear warm-up of 200 steps. We performed training for 10 epochs, saving each epoch and selecting the best one on the validation set.

The training stage in a distributed setting takes approximately 2.38 minutes per epoch (+0.05 for the \emph{avg. embedding} modification) for the TC task. For the SI task, it takes 6.55 minutes per epoch for RoBERTa (+1.27 for CRF), and 6.75 minutes per epoch for XLNet (+1.08 for CRF).

\section{Experiments and Results}\label{sec:6}

\subsection{Token Classification}

We experimented with BIOES, BIO, and IO encodings, and we found that BIO performed  best, both when using CRF and without it. Thus, we used the BIO encoding in our experiments. We further observed a much better recall with minor loss in precision for our ensemble with span merging.

A comparison of the described approaches for the SI and the KI tasks is presented in Table~\ref{table_1}. Although the sequential predictions of the models are generally consistent, adding a CRF layer on top improves the results. Manual analysis of the output for the SI task has revealed that about 3.5\% of the predicted tags were illegal sequences, e.g.,~an ``\texttt{I-PROP}'' tag following an ``\texttt{O}'' tag.

\bgroup
\begin{table}[tbh]
\begin{center}
\small
\begin{tabular}{l l l}
\toprule
\multicolumn{3}{c}{\bf Technique Classification} \\
\midrule
\bf Approach & \bf RoBERTa & \bf XLNet \\
\midrule
Baseline & $62.75$ & $58.23$  \\ 
\hspace{10pt}+ length &  $63.50_{\bf\,\uparrow 0.75}$ &
$59.64_{\bf\,\uparrow 1.41}$ \\
\hspace{10pt}+ averaged span embededding &  $62.94_{\bf\,\uparrow 0.19}$ &
$59.64_{\bf\,\uparrow 1.41}$ \\
\hspace{10pt}+ multi-label &  $63.78_{\bf\,\uparrow 1.03}$ &
$59.27_{\bf\,\uparrow 1.04}$ \\
\hspace{10pt}+ gazetteer post-processing &  $62.84_{\bf\,\uparrow 0.10}$ &
$58.33_{\bf\,\uparrow 0.10}$ \\
\hspace{10pt}+ \emph{repetition} post-processing & $66.79_{\bf\,\uparrow 4.04}$ &
$62.46_{\bf\,\uparrow 3.67}$ \\
\hspace{10pt}+ class insertions &  $62.65_{\,\downarrow 0.10}$ &
$ 57.85_{\,\downarrow 0.38}$ \\
\bottomrule
\end{tabular}
\end{center}
\caption{Analysis of the improvements using RoBERTa and XLNet for the TC task on the development set. Shown is micro-F$_1$ score.}
\label{table_2}
\end{table}

\begin{table}[tbh]
\begin{center}
\small
\begin{tabular}{l l l}
\toprule
\multicolumn{3}{c}{\bf Keyphrase Classification} \\
\midrule 
\bf Approach & \bf RoBERTa & \bf XLNet \\
\midrule
Baseline & $77.18$  & $78.50$ \\ 
\hspace{10pt}+ length &  $77.38_{\bf\,\uparrow 0.20}$ & $78.65_{\bf\,\uparrow 0.15}$ \\
\hspace{10pt}+ gazetteer post-processing &  $77.43_{\bf\,\uparrow 0.25}$ & $78.69_{\bf\,\uparrow 0.19}$ \\
\hspace{10pt}+ class insertions &  $77.82_{\bf\,\uparrow 0.64}$ &
$78.69_{\bf\,\uparrow 0.19}$ \\
\bottomrule
\end{tabular}
\end{center}
\caption{Analysis of improvements for the KC task using RoBERTa and XLNet on the development set in the multi-label mode. Shown is micro-F$_1$ score.}
\label{table_kc}
\end{table}
\egroup

Also, we figured out that neither XLNet nor RoBERTa could learn the described rule for quotes and punctuation symbols. Moreover, adding CRF also does not help solve the problem according to the better ``overall'' score in the table. We analyzed the source of these errors. Indeed, there were some annotation errors. However, the vast majority of the errors related to punctuation at the boundaries were actually model errors. E.g.,~in an example like \texttt{<"It is what it is.">}, where the entire text (including the quotation marks) had to be detected, the model would propose sequences like \texttt{<"It is what it is>} or \texttt{<It is what it is.>}. Thus, there is a common problem for all Transformer-based models---lack of consistency for sequential tag predictions.

\subsection{Sequence Classification}

We took models that use separate inputs (span and context) for all experiments, as they yielded better results on the validation set. The results for the customized models are shown in Tables~\ref{table_2} and \ref{table_kc} for the Technique Classification (TC) and the Keyphrase Classification (KC) tasks, respectively. We also studied the impact of the natural multi-label formulation of the TC task (see Table~\ref{table_3}). We can see that all directions of quality changes were the same.

\bgroup
\begin{table}[t!]
\begin{center}
\small
\begin{tabular}{l l l}
\toprule
\multicolumn{3}{c}{\bf Technique Classification} \\
\midrule 
\bf Approach & \bf RoBERTa & \bf XLNet \\
\midrule
Baseline+multi-label & $63.78$  & $59.27$ \\ 
\hspace{10pt}+ length &  $64.72_{\bf\,\uparrow 0.94}$ & $60.68_{\bf\,\uparrow 1.41}$ \\
\hspace{10pt}+ averaged span embededding &  $64.25_{\bf\,\uparrow 0.47}$ & $60.77_{\bf\,\uparrow 1.50}$ \\
\hspace{10pt}+ gazetteer post-processing &  $63.87_{\bf\,\uparrow 0.09}$ & 
$59.36_{\bf\,\uparrow 0.09}$ \\
\hspace{10pt}+ \emph{repetition} post-processing &  $67.54_{\bf\,\uparrow 3.76}$ &
$63.50_{\bf\,\uparrow 4.23}$ \\
\hspace{10pt}+ class insertions &  $63.69_{\,\downarrow 0.09}$ &
$58.89_{\,\downarrow 0.38}$ \\
\bottomrule
\end{tabular}
\end{center}
\caption{Analysis of the improvements for the TC task using RoBERTa and XLNet on the development set in the multi-label mode. Shown is micro-F$_1$ score.}
\label{table_3}
\end{table}
\egroup

\begin{table}[t!]
\begin{center}
\small
\begin{tabular}{l l}
\toprule
\multicolumn{2}{c}{\bf Technique Classification} \\
\midrule 
\bf Approach & \bf F1-score  \\
\midrule
RoBERTa & $62.08$  \\
\hspace{10pt}+ length and averaged span embedding & $62.27_{\bf\,\uparrow 0.19}$ \\ 
\hspace{26pt}+ multi-label correction & $63.50_{\bf\,\uparrow 1.23}$ \\ 
\hspace{42pt}+ class insertions & $63.69_{\bf\,\uparrow 0.19}$ \\ 
\hspace{58pt}+ \emph{repetition} post-processing & $66.89_{\bf\,\uparrow 3.20}$ \\ 
\hspace{74pt}+ gazetteer post-processing & $67.07_{\bf\,\uparrow 0.18}$ \\
\bottomrule
\end{tabular}
\end{center}
\caption{An incremental analysis of the proposed approach for the TC task on the development set.}
\label{table_4}
\end{table}

Although positional embeddings are used in BERT-like models, our experiments showed that they are not enough to model the length of the span. Indeed, the results for systems that explicitly use length improved both for RoBERTa and for XLNet, for both tasks.

According to the source implementation of RoBERTa, XLNet, and other similar models, only the \texttt{[CLS]} token embedding is used for sequence classification. However, in the TC task, it turned out that the remaining tokens can also be useful, as in the averaging approach.

Moreover, the use of knowledge from the training set through post-processing with a gazetteer consistently improved the results for both models. Yet, it can also introduce errors since it ignores context. That is why we did not set 100\% probabilities for the corrected classes.
 
As for the sequential consistency of labels, the systems produced output with unacceptable nesting of spans of incompatible classes. Thus, correcting such cases can also have a positive impact (see Table~\ref{table_kc}). However, a correct nesting does not guarantee correct final markup, since we only post-process predictions. Better results can be achieved if the model tries to learn this as part of training.

The tables show that the highest quality increase for the TC task was achieved by correcting the \emph{repetition} class. This is because this class is very frequent, but it often requires considering a larger context.

We also examined the impact of each modification on RoBERTa for the TC task, applying an incremental analysis on the development set (Table~\ref{table_4}). We can see that our proposed modifications are compatible and can be used together.

Finally, note that while better pre-training could make some of the discussed problems less severe, it is still true that certain limitations are more ``theoretical'' and that they would not be resolved by simple pre-training. For example, there is nothing in the Transformer architecture that would allow it to model the segment length, etc.

\section{Discussion}
\label{sec:7}

Below we describe a desiderata to add to the Transformer in order to increase its expressiveness, which could guide the design of the next generation of general Transformer architectures.

\paragraph{Length} We have seen that length is important for the sequence labeling task. However, it would be important for a number of other NLP tasks, e.g.,~in seq2seq models. For example, in Neural Machine Translation, if we have an input sentence of length 20, it might be bad to generate a translation of length 2 or of length 200. Similarly, in abstractive neural text summarization, we might want to be able to inform the model about the expected target length of the summary: should it be 10 words long? 100-word long?

\paragraph{External Knowledge} Gazetteers are an important source of external knowledge, and it is important to have a mechanism to incorporate such knowledge. A promising idea in this direction is KnowBERT \cite{Peters_2019}, which injects Wikipedia knowledge when pre-training BERT.

\paragraph{Global Consistency} For structure prediction tasks, such as sequence segmentation and labeling, e.g.,~named entity recognition, shallow parsing, and relation extraction, it is important to model the dependency between the output labels. This can be done by adding a CRF layer on top of BERT, but it would be nice to have this as part of the general model.
More generally, for many text generation tasks, it is essential to encourage the global consistency of the output text, e.g.,~to avoid repetitions. This is important for machine translation, text summarization, chat bots, dialog systems, etc.

\paragraph{Symbolic vs. Distributed Representation} Transformers are inherently based on distributed representations for words and tokens. This can have limitations, e.g.,~we have seen that BERT cannot pay attention to specific symbols in the input such as specific punctuation symbols like quotation marks. Having a hybrid symbolic-distributed representation might help address these kinds of limitations.
It might also make it easier to model external knowledge, e.g.,~in the form of gazetteers.

\section{Conclusion and Future Work}
\label{sec:8}

We have shed light on some important theoretical limitations of pre-trained BERT-style models that are inherent in the general Transformer architecture.
In particular, we demonstrated on two different tasks---one on segmentation, and one on segment labeling---and four datasets that these limitations are indeed harmful and that addressing them, even in some very simple and na\"{i}ve ways, can yield sizable improvements over vanilla BERT, RoBERTa, and XLNet models. Then, we offered a more general discussion on desiderata for future additions to the Transformer architecture in order to increase its expressiveness, which we hope could help in the design of the next generation of deep NLP architectures.
    
In future work, we plan to analyze more BERT-style architectures, especially such requiring text generation, as here we did not touch the generation component of the Transformer. 
We further want to experiment with a pre-formulation of the task as span enumeration instead of sequence labeling with BIO tags.
Moreover, we plan to explore a wider range of NLP problems, again with a focus on such involving text generation, e.g.,~machine translation, text summarization, and dialog systems.

\section*{Acknowledgments}
Anton Chernyavskiy and Dmitry Ilvovsky performed this research within the framework of the HSE University Basic Research Program.

Preslav Nakov contributed as part of the Tanbih mega-project (\url{http://tanbih.qcri.org/}), which is developed at the Qatar Computing Research Institute, HBKU, and aims to limit the impact of ``fake news,'' propaganda, and media bias by making users aware of what they are reading.

%
%
%

\begin{thebibliography}{10}
\bibitem{arkhipov-etal-2019-tuning}
Arkhipov, M., Trofimova, M., Kuratov, Y., Sorokin, A.: Tuning multilingual transformers for language-specific named entity recognition. In: Proceedings of the 7th Workshop on Balto-Slavic Natural Language Processing (BSNLP'19). pp. 89-93. Florence, Italy (2019)

\bibitem{augenstein-etal-2017-semeval}
Augenstein, I., Das, M., Riedel, S., Vikraman, L., McCallum, A.: SemEval 2017 task 10: ScienceIE - extracting keyphrases and relations from scientific publications. In: Proceedings of the 11th International Workshop on Semantic Evaluation (SemEval'17). pp. 546-555.  Vancouver, Canada (2017)

\bibitem{Beltagy2020Longformer}
Beltagy, I., Peters, M.E., Cohan, A.: Longformer: The long-document transformer. In: ArXiv (2020)

\bibitem{DBLP:journals/corr/abs-2009-14794}
Choromanski, K., Likhosherstov, V., Dohan, D., Song, X., Gane, A., Sarl{\'o}s, T., Hawkins, P., Davis, J., Mohiuddin, A., Kaiser, L., Belanger, D., Colwell, L., Weller, A.: Rethinking attention with performers. In: Proceedings of the 9th International Conference on Learning Representations (ICLR'21).  (2021)

\bibitem{DBLP:journals/corr/abs-1906-04341}
Clark, K., Khandelwal, U., Levy, O., Manning, C.D.: What does BERT look at? An analysis of BERT's attention. ArXiv (2019)

\bibitem{DaSanMartinoSemeval20task11}
Da San Martino, G., Barr{\'o}n-Cede{\~n}o, A., Wachsmuth, H., Petrov, R., Nakov, P.: SemEval-2020 task 11: Detection of propaganda techniques in news articles. In: Proceedings of the 14th International Workshop on Semantic Evaluation (SemEval'20), Barcelona, Spain (2020)

\bibitem{da-san-martino-etal-2019-fine}
Da San Martino, G., Yu, S., Barr{\'o}n-Cede{\~n}o, A., Petrov, R., Nakov, P.: Fine-grained analysis of propaganda in news article. In: Proceedings of the 2019 Conference on Empirical Methods in Natural Language Processing and the 9th International Joint Conference on Natural Language Processing (EMNLP-IJCNLP'19). pp. 5636-5646. Hong Kong, China (2019)

\bibitem{Dai_2019}
Dai, Z., Yang, Z., Yang, Y., Carbonell, J., Le, Q., Salakhutdinov, R.: TransformerXL: Attentive language models beyond a fixed-length context. Proceedings of the 57th Annual Meeting of the Association for Computational Linguistics (ACL'19). pp. 2978-2988. Florence, Italy (2019)

\bibitem{devlin-etal-2019-bert}
Devlin, J., Chang, M.W., Lee, K., Toutanova, K.: BERT: Pre-training of deep bidirectional transformers for language understanding. In: Proceedings of the 2019 Conference of the North American Chapter of the Association for Computational Linguistics: Human Language Technologies (NAACL-HLT'19). pp. 4171-4186. Minneapolis, MN, USA (2019)

\bibitem{durrani-etal-2019-one}
Durrani, N., Dalvi, F., Sajjad, H., Belinkov, Y., Nakov, P.: One size does not fit all: Comparing NMT representations of different granularities. In: Proceedings of the 2019 Conference of the North American Chapter of the Association for Computational Linguistics: Human Language Technologies (NAACL-HLT'19). pp. 1504-1516. Minneapolis, MN, USA (2019)

\bibitem{Ettinger_2020}
Ettinger, A.: What BERT is not: Lessons from a new suite of psycholinguistic diagnostics for language models. Transactions of the Association for Computational Linguistics \textbf{8}, 34-48 (2020)

\bibitem{goldberg2019assessing}
Goldberg, Y.: Assessing bert’s syntactic abilities (2019)

\bibitem{jawahar-etal-2019-bert}
Jawahar, G., Sagot, B., Seddah, D.: What does BERT learn about the structure of language? In: Proceedings of the 57th Annual Meeting of the Association for Computational Linguistics (ACL'19). pp. 3651-3657. Florence, Italy (2019)

\bibitem{jin2019bert}
Jin, D., Jin, Z., Zhou, J.T., Szolovits, P.: Is BERT really robust? A strong baseline for natural language attack on text classification and entailment. In: Proceedings of the 34th Conference on Artificial Intelligence (AAAI'20). pp. 8018-8025 (2019)

\bibitem{DBLP:journals/corr/abs-2006-16236}
Katharopoulos, A., Vyas, A., Pappas, N., Fleuret, F.: Transformers are RNNs: Fast autoregressive transformers with linear attention. In: Proceedings of the 37th International Conference on Machine Learning (ICML'20). pp. 5156-5165 (2020)

\bibitem{kovaleva-etal-2019-revealing}
Kovaleva, O., Romanov, A., Rogers, A., Rumshisky, A.: Revealing the dark secrets of BERT. In: Proceedings of the Conference on Empirical Methods in Natural Language Processing and the International Joint Conference on Natural Language Processing (EMNLP-IJCNLP'19). pp. 4365-4374. Hong Kong, China (2019)

\bibitem{10.5555/645530.655813}
Lafferty, J.D., McCallum, A., Pereira, F.C.N.: Conditional random fields: Probabilistic models for segmenting and labeling sequence data. In: Proceedings of the Eighteenth International Conference on Machine Learning (ICML'01). pp. 282-289. Williamstown, MA, USA (2001)

\bibitem{lan2019albert}
Lan, Z., Chen, M., Goodman, S., Gimpel, K., Sharma, P., Soricut, R.: ALBERT: A lite BERT for self-supervised learning of language representations. In: ArXiv (2019)

\bibitem{Liu_2019}
Liu, N.F., Gardner, M., Belinkov, Y., Peters, M.E., Smith, N.A.: Linguistic knowledge and transferability of contextual representations. In: Proceedings of the 2019 Conference of the North American Chapter of the Association for Computational Linguistics: Human Language Technologies (NAACL-HLT'19). pp. 1073-1094. Minneapolis, MN, USA (2019)

\bibitem{DBLP:journals/corr/abs-1907-11692}
Liu, Y., Ott, M., Goyal, N., Du, J., Joshi, M., Chen, D., Levy, O., Lewis, M., Zettlemoyer, L., Stoyanov, V.: RoBERTa: A robustly optimized BERT pretraining approach. In: ArXiv (2019)

\bibitem{peters-etal-2018-deep}
Peters, M., Neumann, M., Iyyer, M., Gardner, M., Clark, C., Lee, K., Zettlemoyer, L.: Deep contextualized word representations. In: Proceedings of the 2018 Conference of the North American Chapter of the Association for Computational Linguistics: Human Language Technologies (NAACL-HLT'18). pp. 2227-2237. New Orleans, LA, USA (2018)

\bibitem{Peters_2019}
Peters, M.E., Neumann, M., Logan, R., Schwartz, R., Joshi, V., Singh, S., Smith, N.A.: Knowledge enhanced contextual word representations. In: Proceedings of the 2019 Conference on Empirical Methods in Natural Language Processing and the 9th International Joint Conference on Natural Language Processing (EMNLP-IJCNLP'19). pp. 43-54. Hong Kong, China. (2019)

\bibitem{Popel_2018}
Popel, M., Bojar, O.: Training tips for the transformer model. The Prague Bulletin of Mathematical Linguistics \textbf{110}(1), 43-70 (2018)

\bibitem{porter1980algorithm}
Porter, M.F.: An algorithm for suffix stripping. Program \textbf{14}(3), 130-137 (1980)

\bibitem{Ratinov2009DesignCA}
Ratinov, L.A., Roth, D.: Design challenges and misconceptions in named entity recognition. In: Proceedings of the Thirteenth Conference on Computational Natural Language Learning (CoNLL'09). pp. 147-155. Boulder, CO, USA. (2009)

\bibitem{rogers2020primer}
Rogers, A., Kovaleva, O., Rumshisky, A.: A Primer in BERTology: What We Know About How BERT Works. Trans. Assoc. Comput. Linguistics 8: 842-866 (2020)

\bibitem{sanh2019distilbert}
Sanh, V., Debut, L., Chaumond, J., Wolf, T.: DistilBERT, a distilled version of BERT: smaller, faster, cheaper and lighter. In: ArXiv (2019)

\bibitem{souza2019portuguese}
Souza, F., Nogueira, R., Lotufo, R.: Portuguese named entity recognition using BERT-CRF. In: Arxiv (2019)

\bibitem{sun2020advbert}
Sun, L., Hashimoto, K., Yin, W., Asai, A., Li, J., Yu, P., Xiong, C.: Adv-BERT: BERT is not robust on misspellings! generating nature adversarial samples on BERT. In: Arxiv (2020)

\bibitem{tenney2019learn}
Tenney, I., Xia, P., Chen, B., Wang, A., Poliak, A., McCoy, R.T., Kim, N., Durme, B.V., Bowman, S.R., Das, D., Pavlick, E.: What do you learn from context? Probing for sentence structure in contextualized word representations. In: Arxiv (2019)

\bibitem{DBLP:journals/corr/VaswaniSPUJGKP17}
Vaswani, A., Shazeer, N., Parmar, N., Uszkoreit, J., Jones, L., Gomez, A.N., Kaiser, L., Polosukhin, I.: Attention is all you need. In: Arxiv (2017)

\bibitem{Wallace_2019}
Wallace, E., Wang, Y., Li, S., Singh, S., Gardner, M.: Do NLP models know numbers? Probing numeracy in embeddings. In: Proceedings of the 2019 Conference on Empirical Methods in Natural Language Processing and the 9th International Joint Conference on Natural Language Processing (EMNLP-IJCNLP'19). pp. 5307-5315. Hong Kong, China (2019)

\bibitem{DBLP:journals/corr/abs-2006-04768}
Wang, S., Li, B.Z., Khabsa, M., Fang, H., Ma, H.: Linformer: Self-attention with linear complexity. In: Arxiv (2020)

\bibitem{NIPS2019_8812}
Yang, Z., Dai, Z., Yang, Y., Carbonell, J., Salakhutdinov, R.R., Le, Q.V.: XLNet: Generalized autoregressive pretraining for language understanding. In: Proceedings of the Annual Conference on Neural Information Processing Systems (NeurIPS'19), pp. 5753-5763. (2019)

\bibitem{DBLP:journals/corr/abs-2007-14062}
Zaheer, M., Guruganesh, G., Dubey, A., Ainslie, J., Alberti, C., Onta{\~{n}}{\'{o}}n, S., Pham, P., Ravula, A., Wang, Q., Yang, L., Ahmed, A.: Big bird: Transformers for longer sequences. In: Proceedings of the Annual Conference on Neural Information Processing Systems (NeurIPS'20). (2020)

\end{thebibliography}

\end{document}